\def\cvprPaperID{5534}
\def\confName{CVPR}
\def\confYear{2023}

\def\paperTitle{Lift3D: Synthesize 3D Training Data\protect\\ by Lifting 2D GAN to 3D Generative Radiance Field}

\def\authorBlock{
    Leheng Li\textsuperscript{1}\thanks{Work done during an internship at NIO Autonomous Driving.} \quad
    Qing Lian\textsuperscript{2} \quad
    Luozhou Wang\textsuperscript{1} \quad
    Ningning Ma\textsuperscript{3} \quad
    Ying-Cong Chen\textsuperscript{1,2}\thanks{Corresponding author.} \\
    \textsuperscript{1}HKUST(GZ) \quad
    \textsuperscript{2}HKUST \quad
    \textsuperscript{3}NIO Autonomous Driving \\
}
\newif\ifreview 
\newif\ifarxiv \newcommand{\arxiv}{\arxivtrue}
\newif\ifcamera 
\newif\ifrebuttal 

\arxiv

\pdfoutput=1
\documentclass[10pt,twocolumn,letterpaper]{article}
\ifreview \usepackage[review]{cvpr} \fi
\ifarxiv \usepackage[pagenumbers]{cvpr} \fi
\ifrebuttal \usepackage[rebuttal]{cvpr} \fi
\ifcamera \usepackage{cvpr} \fi

\usepackage{graphicx}
\usepackage{amsmath}
\usepackage{amssymb}
\usepackage{booktabs}

\usepackage{amssymb}
\usepackage{pifont}

\usepackage{gensymb}

\usepackage[accsupp]{axessibility} 

\usepackage{times}
\usepackage{microtype}
\usepackage{epsfig}
\usepackage[table,xcdraw]{xcolor}
\usepackage{caption}
\usepackage{float}
\usepackage{placeins}
\usepackage{color, colortbl}
\usepackage{stfloats}
\usepackage{enumitem}
\usepackage{tabularx}
\usepackage{xstring}
\usepackage{multirow}
\usepackage{xspace}
\usepackage{url}
\usepackage{subcaption}
\usepackage{xcolor}
\usepackage[hang,flushmargin]{footmisc}

\usepackage{wrapfig}

\ifcamera \usepackage[accsupp]{axessibility} \fi

% ...

\newcommand{\myparagraph}[1]{\vspace{3pt}\noindent\textbf{#1}}
\newcommand{\ray}{\mathbf{r}}
\newcommand{\timenear}{t_n}
\newcommand{\timefar}{t_f}
\newcommand{\absrp}{\sigma}
\newcommand{\Ctrue}{C(\ray)}
\newcommand{\expo}[1]{\exp\left(#1\right)}

\definecolor{col1}{RGB}{232, 161, 148}
\definecolor{col2}{RGB}{148, 187, 232}

\newcommand{\cmark}{\ding{51}}%

\ifarxiv  \fi

\newcommand{\R}[1]{{%
    \textbf{%
        \ifstrequal{#1}{1}{\textcolor{red}{R#1}}{%
        \ifstrequal{#1}{2}{\textcolor{blue}{R#1}}{%
        \ifstrequal{#1}{3}{\textcolor{magenta}{R#1}}{%
        \ifstrequal{#1}{4}{\textcolor{teal}{R#1}}{%
                           \textcolor{cyan}{R#1}%
        }}}}%
    }%
}}

\usepackage{xr-hyper}

\makeatletter
\newcommand*{\addFileDependency}[1]{
  \typeout{(#1)}
  \@addtofilelist{#1}
  \IfFileExists{#1}{}{\typeout{No file #1.}}
}

\makeatother

\usepackage[pagebackref,breaklinks,colorlinks]{hyperref}
\usepackage[capitalize]{cleveref}
\crefname{section}{Sec.}{Secs.}
\crefname{table}{Table}{Tables}
\crefname{figure}{Fig.}{Figs.}

\begin{document}
\title{\paperTitle}
\author{\authorBlock}
\maketitle

\begin{abstract}

This work explores the use of 3D generative models to synthesize training data for 3D vision tasks. The key requirements of the generative models are that the generated data should be photorealistic to match the real-world scenarios, and the corresponding 3D attributes should be aligned with given sampling labels. However, we find that the recent NeRF-based 3D GANs hardly meet the above requirements due to their designed generation pipeline and the lack of explicit 3D supervision. In this work, we propose Lift3D, an inverted 2D-to-3D generation framework to achieve the data generation objectives. Lift3D has several merits compared to prior methods: (1) Unlike previous 3D GANs that the output resolution is fixed after training, Lift3D can generalize to any camera intrinsic with higher resolution and photorealistic output. (2) By lifting well-disentangled 2D GAN to 3D object NeRF, Lift3D provides explicit 3D information of generated objects, thus offering accurate 3D annotations for downstream tasks. We evaluate the effectiveness of our framework by augmenting autonomous driving datasets. Experimental results demonstrate that our data generation framework can effectively improve the performance of 3D object detectors. Project page: \texttt{\href{https://len-li.github.io/lift3d-web/}{len-li.github.io/lift3d-web}}.

\end{abstract}
\section{Introduction}
\label{sec:intro}

\begin{figure}[t!]
    \centering
    \includegraphics[width = 0.48\textwidth, trim = 0 0 0 0, clip]{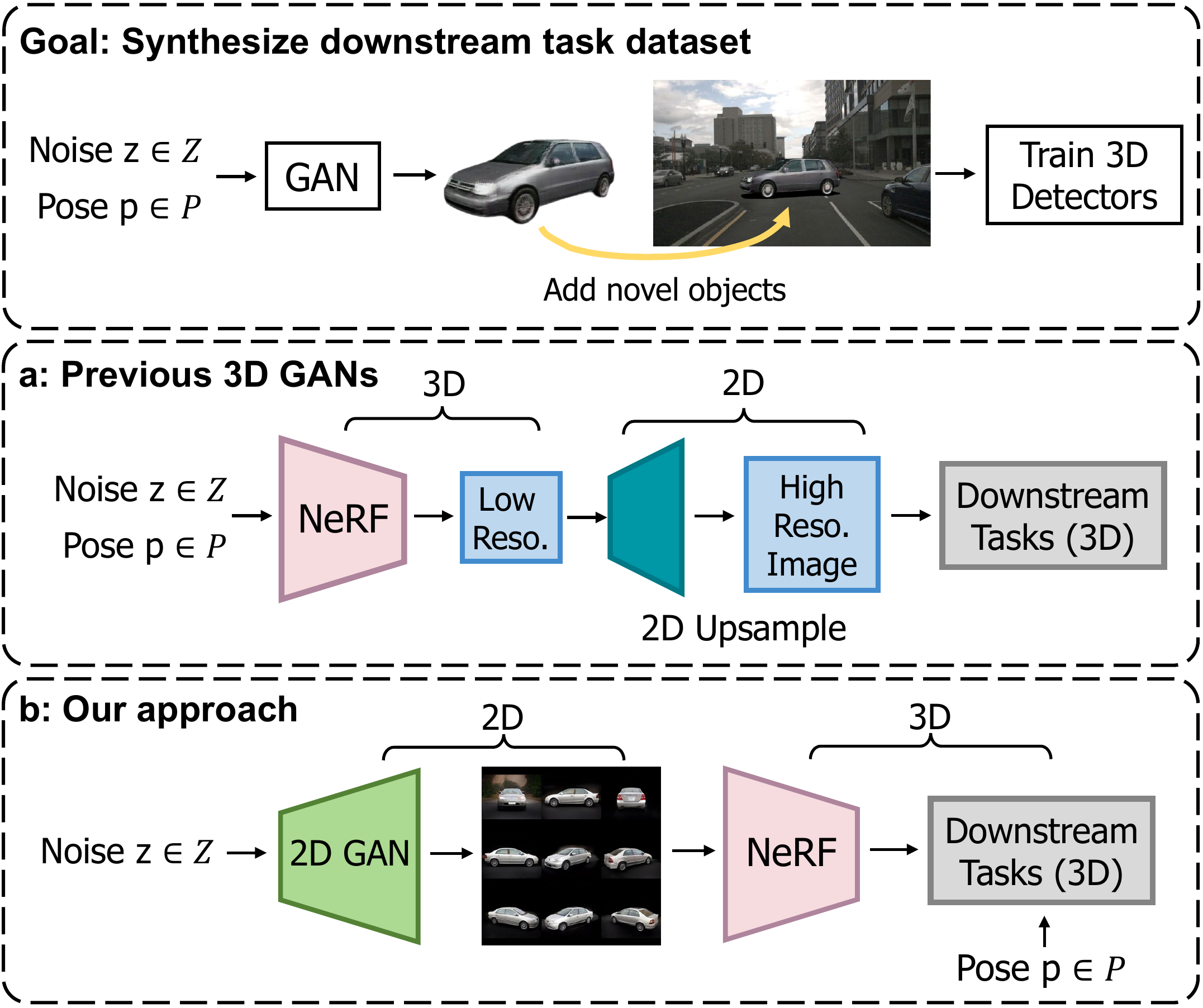}
    \caption{Our goal is to generate novel objects and use them to augment existing datasets. \textbf{(a)} Previous 3D GANs (\textit{e.g.},~\cite{Niemeyer2020GIRAFFE, xue2022giraffehd}) rely on a 2D upsampler to ease the training of 3D generative radiance field, while struggle a trade-off between high-resolution synthesis and 3D consistency. \textbf{(b)} Our 2D-to-3D lifting process disentangles the 3D generation from generative image synthesis, leading to arbitrary rendering resolution and object pose sampling for downstream tasks.}
    \label{fig:net_compare}
\end{figure}




It is well known that the training of current deep learning models requires a large amount of labeled data. However, collecting and labeling the training data is often expensive and time-consuming. This problem is especially critical when the data is hard to annotate.  For example, it is difficult for humans to annotate 3D bounding boxes using a 2D image due to the inherent ill-posed 3D-2D projection (3D bounding boxes are usually annotated using LiDAR point clouds).





To alleviate this problem, a promising direction is to use synthetic data to train our models. For example, data generation that can be conveniently performed using 3D graphics engines offers incredible convenience for visual perception tasks. Several such simulated datasets have been created in recent years~\cite{Weng2020_AIODrive, richter2016playing, gaidon2016virtual, cabon2020virtual, Richter_2017}. These datasets have been used successfully to train networks for perception tasks such as semantic segmentation and object detection. However, these datasets are expensive to generate, requiring specialists to model specific objects and environments in detail. Such datasets also tend to have a large domain gap from real-world ones.

With the development of Generative Adversarial Networks (GAN)~\cite{goodfellow2014generative}, researchers have paid increasing attention to utilize GANs to replace graphics engines for synthesizing training data. For example, BigDatasetGAN~\cite{li2022bigdatasetgan} utilizes condition GAN to generate classification datasets via conditioning the generation process on the category labels. SSOD~\cite{mustikovela2021self} designs a GAN-based generator to synthesize images with 2D bounding boxes annotation for the object detection task. In this paper, we explore the use of 3D GANs to synthesize datasets with 3D-related annotations, which is valuable but rarely explored.

Neural radiance field (NeRF)~\cite{mildenhall2020nerf} based 3D GANs~\cite{Niemeyer2020GIRAFFE, chan2020pi}, which display photorealistic synthesis and 3D controllable property, is a natural choice to synthesize 3D-related training data. 
However, our experimental results show that they struggle to keep high-resolution outputs and geometry-consistency results by relying on a 2D upsampler. Furthermore, the generated images are not well aligned with the given 3D pose, due to the lack of explicit 3D consistency regularization. This misalignment would introduce large label noise in the dataset, limiting the performance in downstream tasks. In addition to these findings, the camera parameters are fixed after training, making them challenging to align the output resolution with arbitrary downstream data.



In this paper, we propose Lift3D, a new paradigm for synthesizing 3D training data by lifting pretrained 2D GAN to 3D generative radiance field. Compared with the 3D GANs that rely on a 2D upsampler, we invert the generation pipeline into 2D-to-3D rather than 3D-to-2D to achieve higher-resolution synthesis. As depicted in Fig.~\ref{fig:net_compare}, we first take advantage of a well-disentangled 2D GAN to generate multi-view images with corresponding pseudo pose annotation. The multi-view images are then lifted to 3D representation with NeRF reconstruction. In particular, by distilling from pretrained 2D GAN, lift3D achieves high-quality synthesis that is comparable to SOTA 2D generative models. By decoupling the 3D generation from generative image synthesis, Lift3D can generate images that are tightly aligned with the sampling label. Finally, getting rid of 2D upsamplers, Lift3D can synthesize images in any resolution by accumulating single-ray evaluation. With these properties, we can leverage the generated objects to augment existing dataset with enhanced quantity and diversity.

To validate the effectiveness of our data generation framework, we conduct experiments on image-based 3D object detection tasks with KITTI~\cite{Geiger2013IJRR} and nuScenes~\cite{nuscene} datasets. Our framework outperforms the best prior data augmentation method~\cite{lian2022copypaste} with significantly better 3D detection accuracy. Furthermore, even without any labeled data, it achieves promising results in an unsupervised manner. Our contributions are summarized as follows:

\begin{itemize}
    \item We provide the first exploration of using 3D GAN to synthesize 3D training data, which opens up a new possibility that adapts NeRF's powerful capabilities of novel view synthesis to benefit downstream tasks in 3D.
    \item To synthesize datasets with high-resolution images and accurate 3D labels, we propose Lift3D, an inverted 2D-to-3D data generation framework that disentangles 3D generation from generative image synthesis.
    \item Our experimental results demonstrate that the synthesized training data can improve image-based 3D detectors across different settings and datasets.
\end{itemize}







\section{Related Work}
\label{sec:related}

\begin{figure}[t!]
    \centering
    \includegraphics[width = 0.48\textwidth, trim = 0 0 0 0, clip]{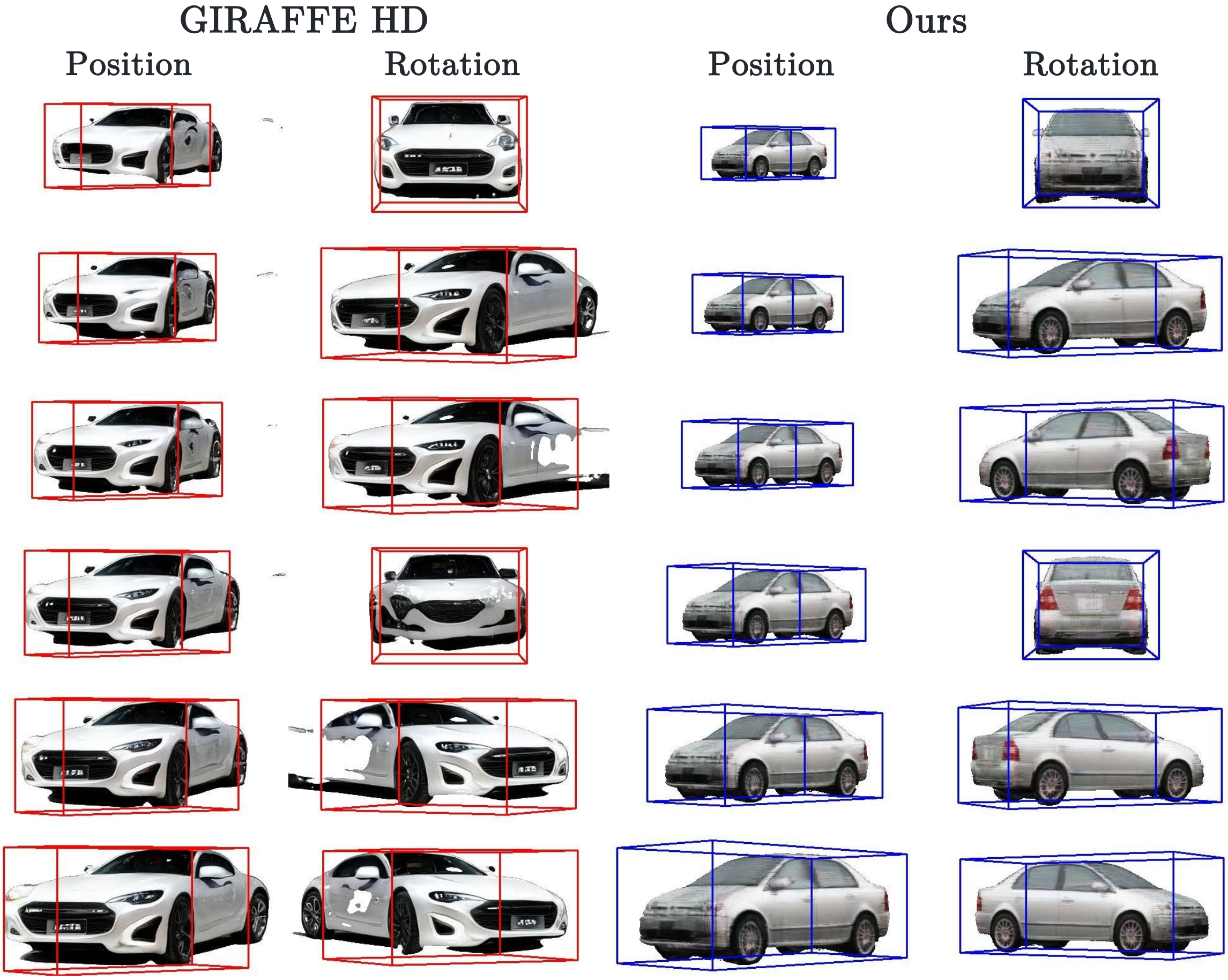}
    \caption{We compare our generation result with GIRAFFE HD~\cite{xue2022giraffehd}. We zoom in or rotate the sampled 3D box to control the generation of models. The rotation of the 3D box introduces artifacts to images generated by GIRAFFE HD. All images are plotted with sampled 3D bounding boxes.}
    \label{fig:compare_gir}
\end{figure}


\myparagraph{Data Generation for Downstream Tasks}
\quad Benefiting from low data acquisition costs, learning from synthesized data is an attractive way to scale up the training data. Several studies like~\cite{richter2016playing, gaidon2016virtual, cabon2020virtual, Alhaija2018IJCV} leverage graphic engines to synthesize training data without human annotation. However, they rely on pre-built 3D assets to mimic the world, which is also a non-negligible effort in the whole pipeline. 

Without any burden of collecting 3D assets, generative models can also be considered as a neural rendering alternative to graphics engines. 
For example, BigDatasetGAN~\cite{li2022bigdatasetgan} generates classification datasets by conditioning the generation process on the class labels. SSOD~\cite{mustikovela2021self} samples dataset with 2D bounding boxes via generative image synthesis. Our method goes further, utilizing 3D GAN to generate training data with 3D annotation, greatly reducing labeling effort in 3D data.

\myparagraph{3D-aware Generative Image Synthesis}
\quad Recently, Generative Adversarial Networks (GANs)~\cite{goodfellow2014generative} have made great progress in generating high-resolution photorealistic 2D images. One natural extension of the 2D GANs is to endow their 3D controllable ability as 2D images are projections of the 3D world. 
To provide the 3D-aware ability, recent work~\cite{chan2020pi, Niemeyer2020GIRAFFE} leverages neural implicit representation (\textit{i.e.}, NeRF~\cite{mildenhall2020nerf}), a 3D rendering module, to represent the generator of GANs. In these methods, the training of NeRF module is via the adversarial discriminator between the generated and real images. Due to a lack of 3D supervision, the 3D consistency between the view angle and object position with generated images is not explicitly regularized, making the objects can not align well with provided 3D information as depicted in Fig.~\ref{fig:compare_gir}. Furthermore, the high computation cost of volumetric rendering in the generation process also makes these methods suffer from scaling issues (\textit{e.g.}, resolution). To alleviate this issue, several methods~\cite{Niemeyer2020GIRAFFE,xue2022giraffehd,Chan2022,orel2022stylesdf,gu2021stylenerf,deng2022gram} choose to leverage a separate 2D upsampler to produce a high-resolution image from a low-resolution feature grid rendered from NeRF (Fig.~\ref{fig:net_compare} (a)). Although this method achieves high-resolution synthesis, the underlying multi-view consistency is non-satisfactory as shown in Fig.~\ref{fig:compare_gir}.



Interpreting the latent representation of GANs~\cite{shen2021closedform,ganspace} has benefited a body of work disentangling various factors of generated objects in a 3D-controllable manner, \textit{e.g.}, viewpoint and shape~\cite{shi2021lifting, StyleGAN3D, liu2022gansvr}. This indicates we can tune 2D GANs into 3D-aware generators by interpreting different factors of disentangled latents. Owing to this powerful latent representations of 2D GANs, we can achieve both high-resolution photorealistic image synthesis and rough 3D controllable property. For example, GANverse3D~\cite{StyleGAN3D} manually annotates the pose labels of StyleGAN2 latents and distills the 3D-aware generation to a mesh-based rendering module. Shi et al.~\cite{shi2021lifting} disentangle and distill the 3D information from StyleGAN2 to 3D-aware face generation. Although they achieve 3D controllability of GAN, they haven't tried to convey the 3D-aware ability to benefit downstream tasks.

\myparagraph{Data Augmentation in Object Detection}
\quad Data augmentation is an effective technique for improving the performance of object detection.  Although several data augmentation methods have yielded impressive gains for 2D tasks, they are hardly adapted to 3D vision tasks due to the violation of geometric relationship (\textit{i.e.}, when manipulating the object in the image domain, it's non-trivial to obtain the corresponding bounding boxes in the 3D world.)  
To alleviate this issue, recent work~\cite{lian2022copypaste, zhang2020multimodality, chen2021geosim} designs geometry-aware copy-paste methods that fix the relationship among the viewing angle, 2D and 3D position to obtain the corresponding 3D ground truth. However, these methods are limited to generating objects with fixed viewing angles and positions. They are also restricted to use pre-collected asset banks. 

In this work, we design a generative-based data generation method for 3D object detection, in which an unlimited amount of objects with arbitrary position and rotation can be generated for training. Based on our method, the diversity of the training data can be effectively enhanced, leading to better coverage of long-tail scenarios in the safety-critical driving domain.

\section{Method}
\label{sec:method}




\subsection{Overview}
\begin{figure*}[!htb]
    \centering
    \includegraphics[width = \textwidth, trim = 0 0 0 0, clip]{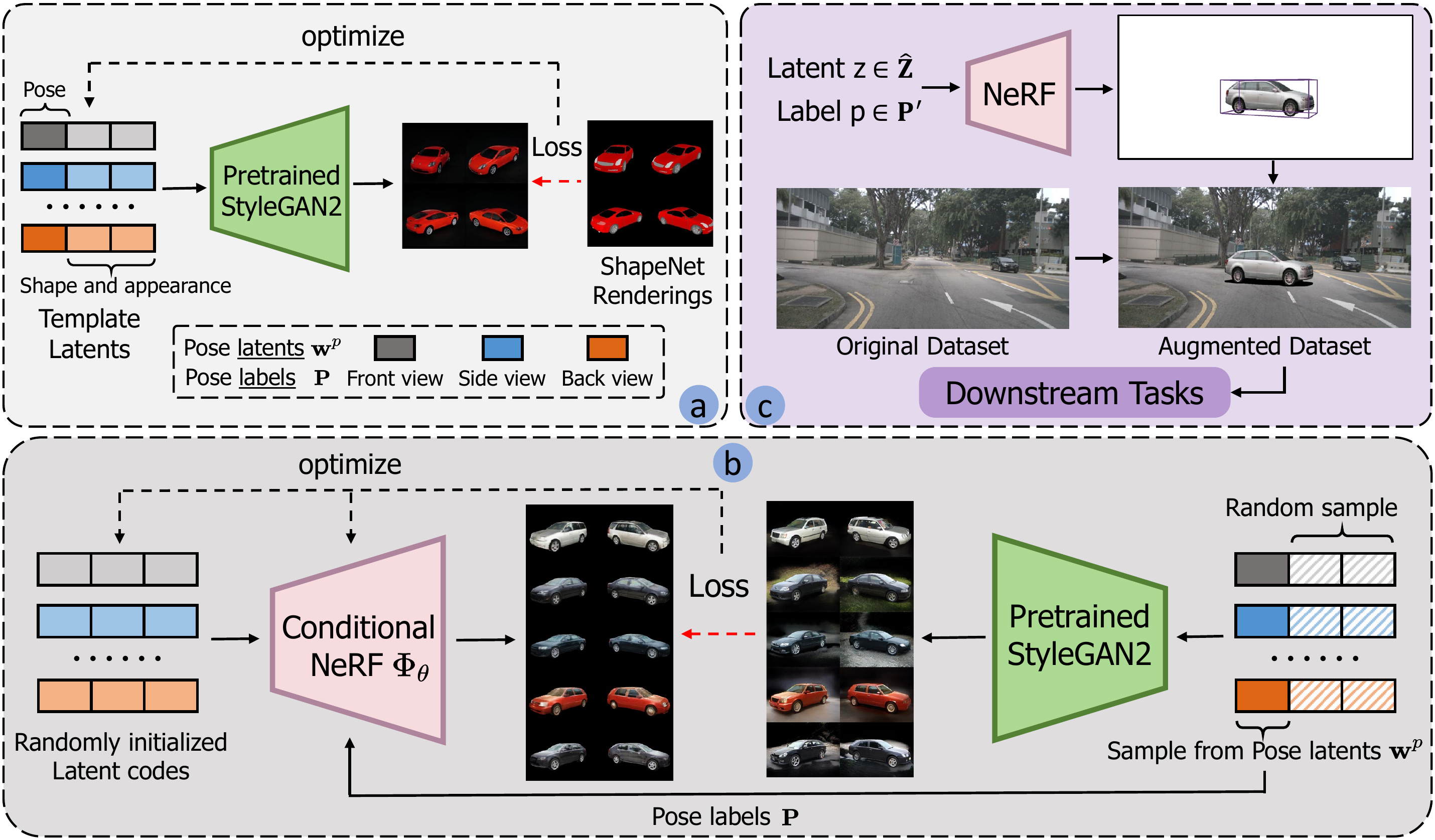}
    \caption{\textbf{Lift3D} is a 3D generation framework that enables training data generation by providing photorealistic synthesis and precise 3D annotation. The framework comprises three key modules: \textbf{(a)} Interpretation of StyleGAN2: We obtain latent-label pairs with the guidance of the ShapeNet renderings. \textbf{(b)} Lifting process: We optimize a shared conditional NeRF and randomly initialized latent codes to lift pretrained StyleGAN2 to 3D object NeRF. \textbf{(c)} Training data generation: By sampling latents $\mathbf{z}$, we can augment the existing dataset by adding novel objects with label $p$.}
    \label{fig:overview}
\end{figure*}

We visualize the generation and augmentation pipelines in Fig.~\ref{fig:overview}. Our method aims to generate novel objects with corresponding 3D bounding boxes annotation and compose them into existing scenes for training. Without loss of generality, we adopt monocular 3D object detection as the downstream task to evaluate the effectiveness of our method. 

\subsection{3D GANs}\label{sec:3dgan}

Recently, the community has witnessed an explosion in the 3D generative radiance field~\cite{Niemeyer2020GIRAFFE,xue2022giraffehd,Chan2022,orel2022stylesdf,gu2021stylenerf,deng2022gram, Zhou2021CIPS3D, xu2021volumegan, zhao_gmpi2022}. However, as revealed in ~\cite{epigraf,gu2021stylenerf}, none of them can preserve strict multi-view consistency, partially on account of the usage of a 2D upsampler and lack of explicit 3D supervision. Fig.~\ref{fig:compare_gir} shows the generation result of GIRAFFE HD~\cite{xue2022giraffehd}, the SOTA 3D generative network on Car. When we render objects in fixed rotation and increasing depth, the objects gradually lose details during zoom-out. In the meantime, during sampling in fixed depth and varying rotation, the images break multi-view consistency as the objects only fit the 3D box in a single view. These unsatisfactory results prevent their application in downstream tasks.





To benefit downstream tasks with 3D generative synthesis, we design a method that \textbf{(i)} is able to generate high-resolution photorealistic images, \textbf{(ii)} its generated images are consistent with given sampling label, \textbf{(iii)} capable of generating images in any wanted distribution (not restrict to object-centric views).


The general idea of our method is to switch the order of generation process to 2D-3D rather than 3D-2D. We first introduce our efficient interpretation of StyleGAN2 that serves for dense multi-view synthesis in Sec.~\ref{sec:disentangle}. Then, we detail the proposed lifting process that provides high-resolution synthesis and accurate 3D annotation in Sec.~\ref{sec:nerfrecon}.








\subsection{StyleGAN as a Proxy 3D Generator}\label{sec:disentangle}

To achieve \textbf{(i)}, we propose to incorporate powerful StyleGAN2~\cite{Karras2020stylegan2} to serve as a proxy 3D generator. Our goal is to leverage pretrained StyleGAN2 to densely sample multi-view images with pose labels. Note that other GANs may also work, but we choose~\cite{Karras2020stylegan2} for its most photorealistic synthesis.


We draw aspiration from GANSpace~\cite{ganspace} and SeFa~\cite{shen2021closedform} to disentangle StyleGAN2. Specifically, StyleGAN2 is a synthesis network that maps a latent code $\mathbf{z} \in \mathbb{R}^{512}$ sampled from a Gaussian distribution $\mathbf{Z}\in\mathcal{N}(0, \mathbf{I})$ into a real image. The latent $\mathbf{z}$ is first transformed to ${\mathbf{w}}^*=(w_1^*, w_2^*, ..., w_{16}^*) \in  \!\mathbb{R}^{16\times 512}$ through 16 affine transformations. Then $\mathbf{{w}^*}$ is serve as conditional information to modulate different layers of the synthesis network. GANSpace~\cite{ganspace} found that different layers in $\mathbf{{w}^*}$ control different image attributes. For example, styles in the early layers adjust the camera viewpoint, while styles in the intermediate and higher layers influence shape, texture, and background. That's to say, if we sample a new latent but keep the early layers unchanged, we generate images of an object with a different shape and appearance but depicted in the same viewpoint. This motivates us to  transfer high-quality StyleGAN2 to 3D-aware generators without retraining 3D GANs.


Following GANSpace~\cite{ganspace}, we empirically set the first 8 layers of latents as \emph{pose} latents  (denoted as $\mathbf{w}^p$), the remaining 8 layers of latents as \emph{shape and appearance} latents. Fig.~\ref{fig:overview}~(a) shows how we obtain latent-label pairs $\mathbf{w}^p$ and $\mathbf{P}$ for the lifting process. Since graphics-based rendered images naturally provide pose labels during rendering, we randomly select a 3D car model from ShapeNet~\cite{chang2015shapenet}, and render the model under 200 different viewpoints $\mathbf{P}$, then use optimization-based GAN inversion method~\cite{xia2022gan} to find the corresponding template latents. Thus, the first 8 layers of template latents are associated with meaningful pose information $\mathbf{P}$ (\textit{e.g.}, the pose latent $\mathbf{w}^p$ in darker gray represents a rendering from the front viewpoint). We next illustrate the usage of these 200 latent-label pairs $\mathbf{w}^p$ and $\mathbf{P}$ in Sec.~\ref{sec:nerfrecon}.







\subsection{Lifting 2D Images to 3D Radiance Field}\label{sec:nerfrecon}

 
Directly applying a 3D GAN in downstream tasks is non-trivial as discussed in Sec.~\ref{sec:3dgan}. In this work, we propose a method that distills 3D knowledge from pretrained StyleGAN2: lifting generated images to 3D representation. As depicted in Fig.~\ref{fig:overview}~(b), we jointly optimize a shared conditional NeRF and latent codes to lift StyleGAN2-generated multi-view images to 3D object NeRF. Detailed network structures are described below.




\myparagraph{Generating 2D Data $(\mathbf{I}, \mathbf{P})$}
\quad As shown in the right side of Fig.~\ref{fig:overview}~(b), we leverage the above disentangled StyleGAN2 to sample images $\mathbf{I}$ with pose label $\mathbf{P}$ for the lifting process. During sampling, the first 8 layers of latents are sampled from the pose latents $\mathbf{w}^p$. The remaining 8 layers of latents are randomly sampled from Gaussian distribution $\mathbf{Z}$ for diverse synthesis.




\myparagraph{Incorporating Priors via Global Optimization $\mathbf{\hat{Z}}, \Phi_{\theta}$}
\quad A naive method of lifting multi-view images to 3D assets is to reconstruct each object with an isolated NeRF. However, due to the randomness of GAN sampling and imperfect GAN interpretation, the generated images and pose labels are noisy, or even sometimes corrupted, hindering the reconstruction of a single object. This method also neglects the inductive bias in a certain category. To address the above issues, we model the lifting process as Generative Latent Optimization (GLO)~\cite{bojanowski2017optimizing}, in which each object is assigned a corresponding latent vector. We optimize a shared NeRF $\Phi_{\theta}(z,P)$ with parameter $\theta$ and randomly initialized latent codes $\mathbf{{z}} \in \mathbf{Z}$ for all objects at the same time, thereby granting NeRF generator to learn the shared shape and appearance prior within the same category. The formula can be written as follows: 
\begin{equation}
\mathbf{\hat{Z}}, \hat{\theta} = \arg\min\limits_{\mathbf{{z}}, \theta} \mathcal{L}(\mathbf{I}, \Phi_{\theta}(\mathbf{{z}}, \mathbf{P})),
\end{equation}
where $\mathbf{\hat{Z}}$ represents the concatenation of all latent codes, $\mathbf{P}$ is the pose label obtained from GAN interpretation, $\mathcal{L}$ is our loss function that will be introduced in Sec.~\ref{sec:modellearn}.




\myparagraph{Efficient 3D Representation $\Phi_{\theta}$}
\quad We represent our NeRF network $\Phi_{\theta}$ with a tri-plane~\cite{Chan2022} representation for efficient feature encoding. Specifically, we follow~\cite{Chan2022, Karras2020stylegan2} and use a 2D convolution neural network to map constant input to three axis-aligned orthogonal feature planes with the condition of latent code $\mathbf{{z}}$. For any sampled 3D point $x\in  \!\mathbb{R}^{3}$ of NeRF, we query its feature vector by projecting it onto each of the three feature planes, retrieving the corresponding three feature vectors via bilinear interpolation, then summing the three as the final feature vector. To further incorporate global information, we then feed the final feature vector to a single layer SIREN-based~\cite{sitzmann2020siren} MLP that conditions on $\mathbf{{z}}$ to output the density and RGB value.


\myparagraph{Volumetric Rendering}
\quad The expected color $C(\mathbf{r})$ of camera ray $\mathbf{r}(t)=\mathbf{o} + t\mathbf{d}$ with nearest and furthest sampling bounds $\timenear$ and $\timefar$ is: 

\begin{equation}
\Ctrue = \int_{\timenear}^{\timefar}T(t)\absrp(\mathbf{r}(t))\mathbf{c}(\mathbf{r}(t),\mathbf{d})dt\,,
\end{equation}

\begin{equation}
\textrm{ where } T(t) = \expo{-\int_{\timenear}^{t}\absrp(\mathbf{r}(s))ds}\,.
\end{equation}
The function $T(t)$ denotes the accumulated transmittance along the ray from $\timenear$ to $t$.

We also predict the binary mask of objects $\mathbf{\hat{M}}$ by the accumulated transmittance of the furthest sampling point to extract the foreground pixels. 


\begin{equation}
    \mathbf{\hat{M}} = \left\{\begin{matrix}
 1& & \textup{$ T(t_f) \geq 0.5 $} \\ 
 0& & \textup{$ T(t_f) < 0.5 $}.
\end{matrix}\right.
\end{equation}


\begin{figure}[t!]
    \centering
    \includegraphics[width = 0.48\textwidth, trim = 0 0 0 0, clip]{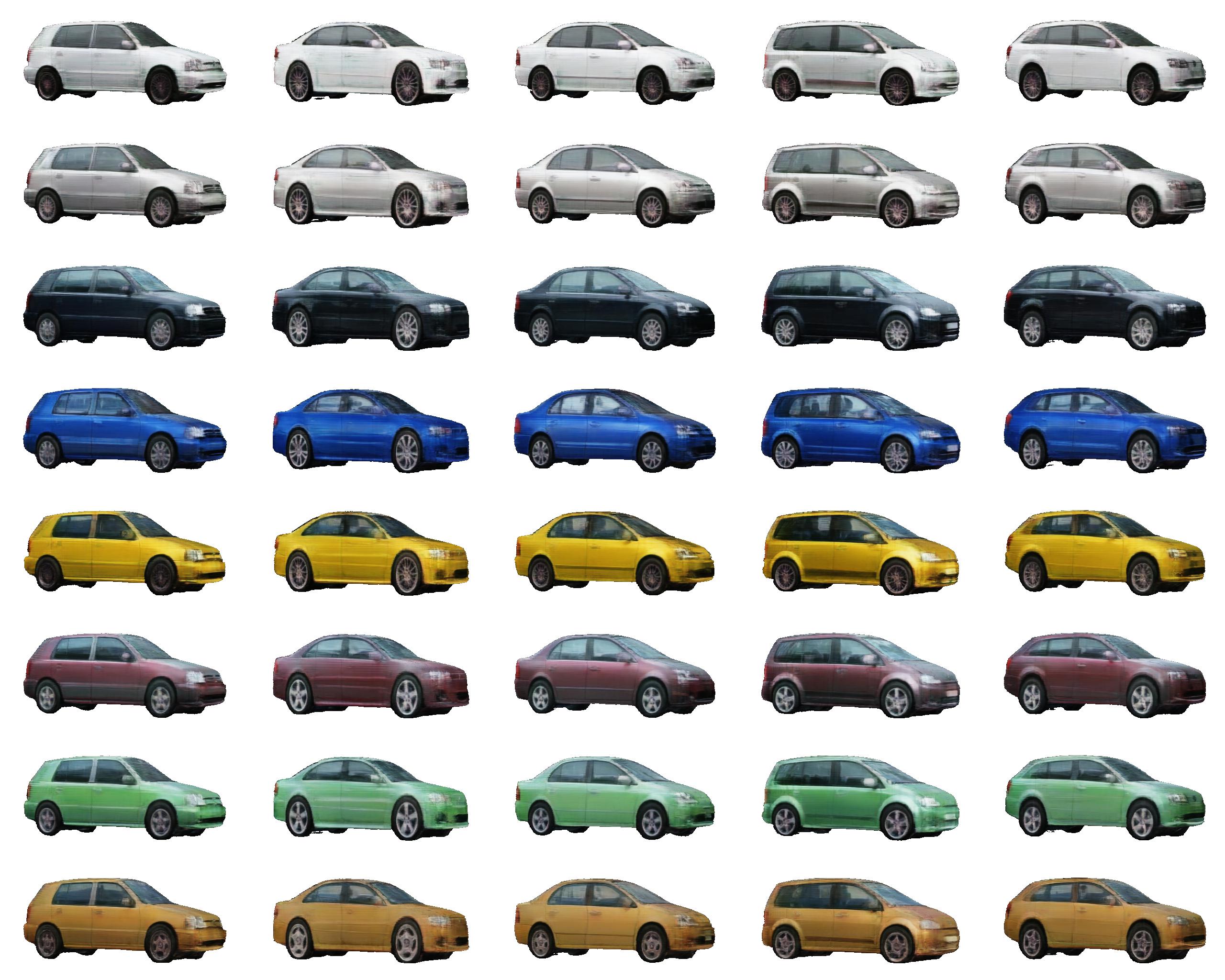}
    \caption{Disentanglement of our NeRF generation. By distilling knowledge from pretrained StyleGAN2, our 3D generation framework directly inherits the well-disentangled property to generate the shape-appearance disentangled images.}
    \label{fig:car_vis}
\end{figure}


\subsection{Sample and Composition}\label{sec:composition}



During sampling, a significant advantage of our method is that: we can synthesize any-resolution images by accumulating single-ray evaluation without relying on upsamplers. Fig.~\ref{fig:overview}~(c) illustrates our sampling procedure.


\myparagraph{Sample Position}
\quad Our goal is to augment the existing dataset by populating novel objects with any wanted pose $\mathbf{P}^{\prime}$. To place objects, a simple way is to uniformly sample positions on road planes, as camera height information can be extracted from calibration. However, this method does not consider relations between objects and environment layout \textit{e.g.}, vehicles don't appear inside a building. To mitigate this problem, we use a pretrained semantic segmentation model HRNet~\cite{SunXLW19} to extract drivable areas in the 2D image plane. Then we convert the perspective view into the bird’s-eye-view (BEV) through Inverse Perspective Mapping (IPM). During sampling, objects that are outside the drivable area will be filtered out. Detailed parametrization of $\mathbf{P}^{\prime}$ can be found in Supp. Material.

\myparagraph{Shadow Generation}
\quad As a non-negligible consequence of lighting interactions, shadows play a vital role in both visual quality~\cite{chen2021geosim} and downstream tasks~\cite{Dijk_2019_ICCV, wang2022neural}. Similar to GeoSim~\cite{chen2021geosim}, we cast a pre-computed shadow map at the bottom of each object. 


\myparagraph{Composition}
\quad  After generating the foreground images, alpha blending is utilized to maintain the visual quality of the object border when composing the foreground and background images. Specifically, we compute the output image as $I_1 \times M + I_2 \times (1 - M)$ where $I_1$ and $I_2$ are the background and foreground images, respectively. We apply a Gaussian filter to foreground mask $M$ to smooth out the edges of the pasted objects.


\subsection{Model Learning}\label{sec:modellearn}

In this section, we detail the learning objective of the lifting process. Given $N$ pairs of multi-view images with pose annotations $\{{I}_{i}, {P}_{i} \}_{i=1}^{N}$, we jointly optimize latents and NeRF generator parameters to minimize the following loss:
\begin{equation}
    \mathcal{L} = \sum_{i=1}^{N} (\mathcal{L}_{RGB}+\lambda_{IoU}\mathcal{L}_{IoU}+\lambda_{perc}\mathcal{L}_{perc}),
\end{equation}
where $\mathcal{L}_{RGB}$ is photometric loss, $\mathcal{L}_{IoU}$ is IoU loss, $\mathcal{L}_{perc}$ is perceptual loss, and $\lambda_*$ are loss weights.   


\myparagraph{Photometric Loss}
\quad In order to optimize both latents and NeRF generator to faithfully reconstruct input objects, we  encourage the pixels of the output image $\Phi_{\theta}(\mathbf{z_i}, {P_i})$ to exactly match the pixels of the input image $I_i$. If both have shape $C \times H \times W$, then the pixel
loss is defined as
\begin{equation}
    \mathcal{L}_{RGB} = \frac{1}{C  H  W}|I_i - \Phi_{\theta}(\mathbf{z_i}, {P_i)}|, 
    \label{eqn:L_rgb_noMap}
\end{equation}
where $|\cdot|$ denotes the $L_1$ loss. $\mathbf{z_i}$ and ${P_i}$ are object latents and pose labels, respectively.




\myparagraph{Perceptual Loss}
\quad We use perceptual loss to preserve high-level feature reconstruction. The loss network \emph{VGG} is the 16-layer VGG network~\cite{vgg_2015} pretrained on ImageNet~\cite{deng2009imagenet}.
\begin{equation}
    \mathcal{L}_{perc} = \frac{1}{C  H  W}|VGG(I_i) - VGG(\hat{I}_i)|. 
    \label{eqn:L_rgb_noMap}
\end{equation}

\myparagraph{IoU Loss}
\quad In addition, we use IoU loss to enforce silhouette consistency. 
\begin{equation}
    \mathcal{L}_{IoU} = 1 - \frac{\hat{M} \cap M^*}{\hat{M} \cup M^*},
    \label{eqn:L_rgb_noMap}
\end{equation}
where $\hat{M}$ is the binary mask predicted by volumetric rendering, $M^*$ is the ground truth binary mask predicted by pretrained segmentation model PointRend~\cite{kirillov2020pointrend}.



\section{Experiment}
\label{sec:Experiment}


In this section, we evaluate the benefits of our 3D generative synthesis as data augmentation for the downstream monocular 3D object detection task on the KITTI~\cite{geiger2012we} and nuScenes datasets~\cite{nuscene}.


\subsection{Experimental Setup}\label{sec:exp_setup}



\myparagraph{KITTI ~\cite{geiger2012we}}
\quad KITTI dataset is one of the most famous autonomous driving datasets, which contains 7,481 training and test frames with 80,256 annotated 3D bounding boxes. Following~\cite{chen20153d, chen2016monocular}, we split the training data into the training and validation subsets and evaluate the effectiveness of our proposed method on the validation set.

\myparagraph{nuScenes ~\cite{nuscene}}
\quad nuScenes is a large-scale autonomous driving dataset. This dataset is collected using 6 surrounded-view cameras that cover the full 360° field of view around the ego-vehicle. Compared with the KITTI dataset, nuScenes dataset contains 7x as many 3D bounding boxes in the urban scenes. 


\begin{figure}[t!]
    \centering
    \includegraphics[width = 0.48\textwidth, trim = 0 0 0 0, clip]{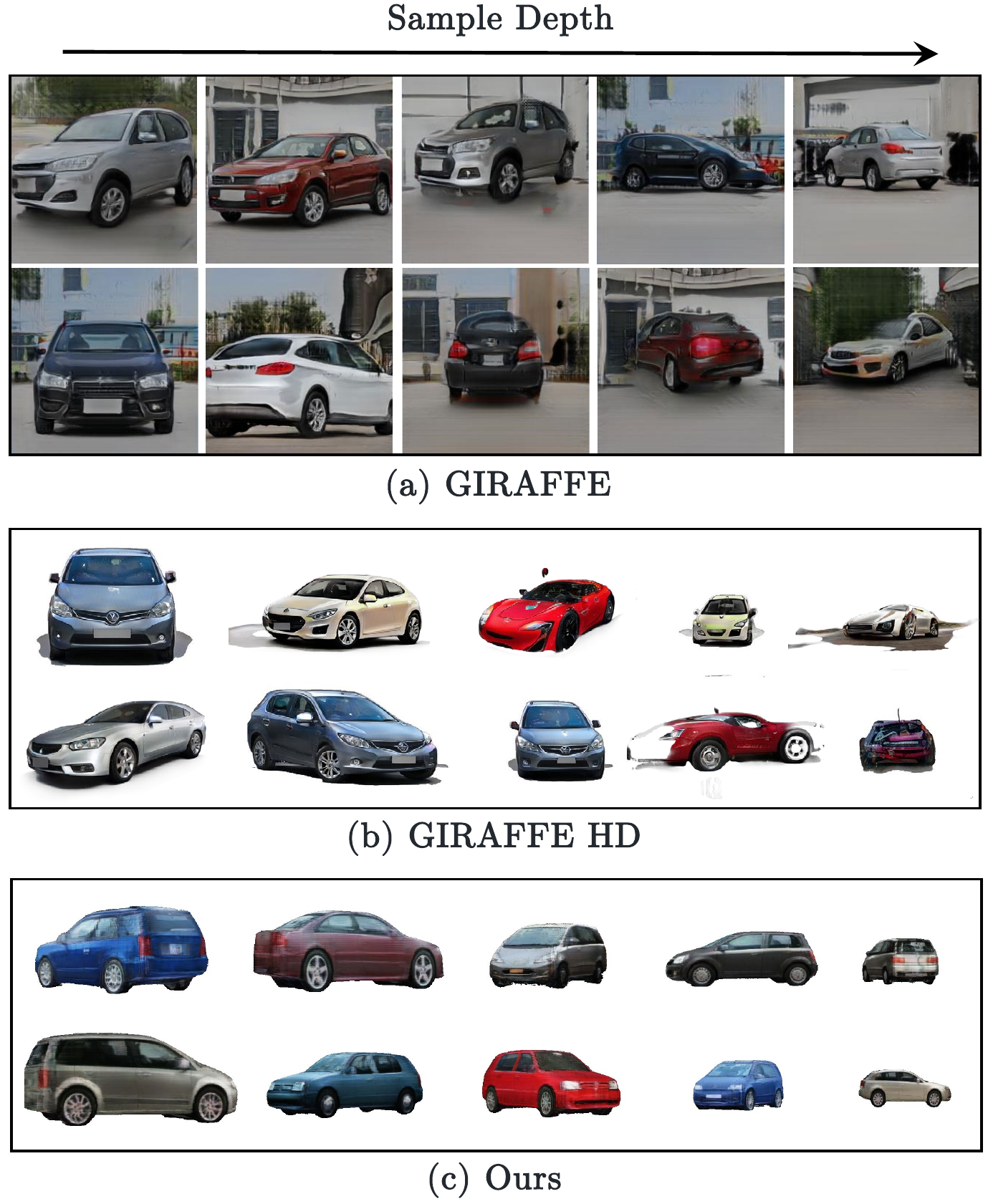}
    \caption{Qualitative image comparisons between GIRAFFE~\cite{Niemeyer2020GIRAFFE}, GIRAFFE HD~\cite{xue2022giraffehd} and Ours. We render objects with randomly sampled rotation while increasing the depth magnitude.}
    \label{fig:compare_g_hd}
\end{figure}


\myparagraph{Evaluation Metrics}
\quad In the KITTI dataset, we follow the official protocol ~\cite{geiger2012we} and adopt the $AP|_{40}$ evaluation metrics on 3D bounding box estimation tasks. The evaluation is conducted separately based on the difficulty levels (Easy, Moderate, and Hard). In the nuScenes dataset, we adopt the provided ~\cite{nuscene} evaluation metrics to evaluate mAP of all classes. We follow a similar setting in ~\cite{wang2022neural}. We use a $10\%$ subset of real data from the nuScenes training set. The subset is sampled scene-wise to mimic label-efficient scenarios.

\myparagraph{Asset Bank Creation}
\quad For StyleGAN2~\cite{Karras2020stylegan2}, we adopt the official model\footnote{\url{https://github.com/NVlabs/stylegan2}} trained on the LSUN Car dataset~\cite{yu2015lsun} to synthesize the multi-view images in an offline manner. We synthesize images with pose label $\mathbf{P}$, which is sampled from the surface of a unit sphere that looks at the origin, with $0-360^{\circ}$ in azimuth and $0-20^{\circ}$ in elevation. We use segmentation network PointRend~\cite{kirillov2020pointrend} to extract binary mask of objects. During inference, images with a confidence lower than 0.5 will be regarded as noisy data to filter out. We created a large object bank of 1000 different vehicles, and each vehicle is assigned 200 multi-view images (note that nothing prevents us from generating a larger amount of data). Then, we conduct the lifting process in the whole asset bank.




\subsection{Comparisons}\label{sec:comparisons}

\begin{figure*}[!htb]
    \centering
    \includegraphics[width = \textwidth, trim = 0 0 0 0, clip]{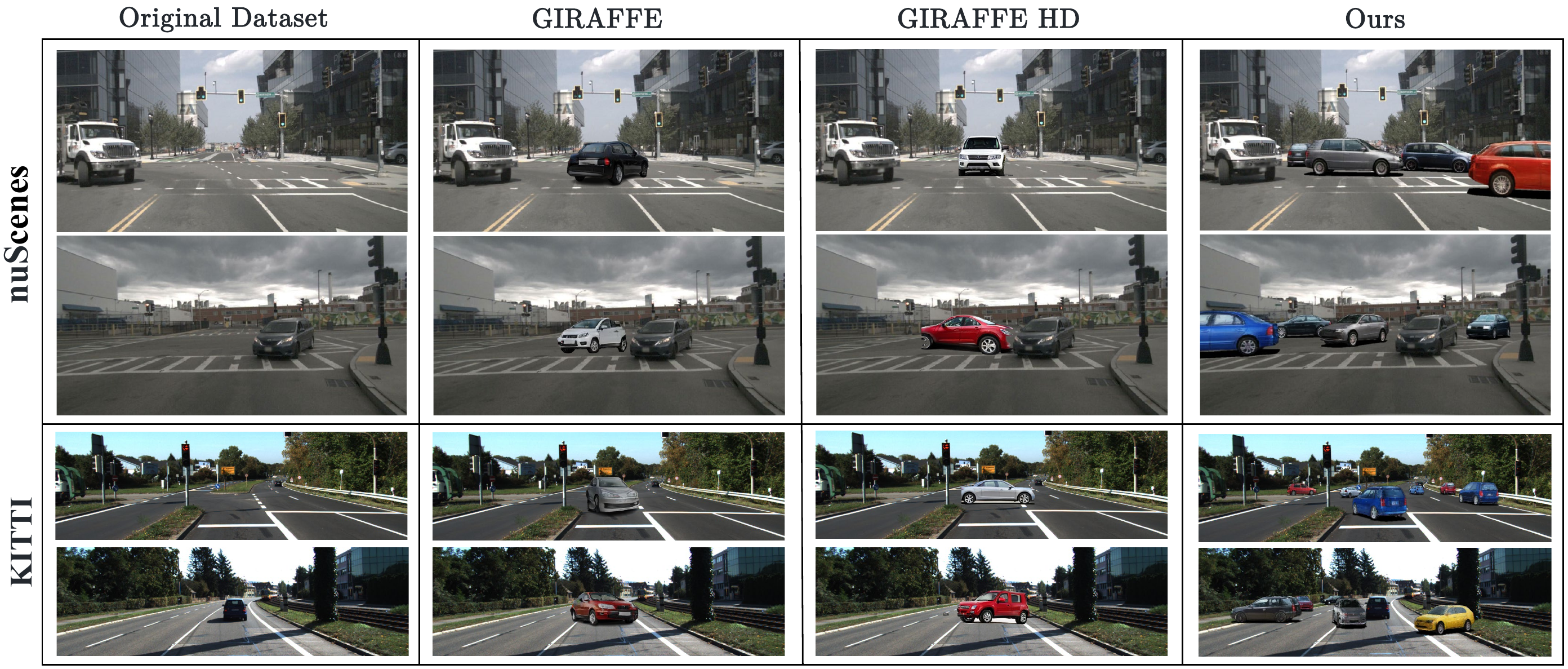}
    \caption{Qualitative comparison of 3D generative synthesis serves as data augmentation on nuScenes and KITTI datasets. Our method is able to generate more realistic shapes and textures, and the resulting objects blend more naturally with the background image.}
    \label{fig:cutpaste_compare}
\end{figure*}

\myparagraph{Qualitative Results}
\quad We compare the visual quality with the existing methods in Fig.~\ref{fig:compare_g_hd}. We utilize the foreground mask in the generative model to filter out the background pixels. Without any deformation of objects or broken parts, our method is significantly more realistic than other methods. Fig.~\ref{fig:cutpaste_compare} further shows the composition result of generated objects and the original datasets, where our method achieves harmonious visual quality without any post-processing.

\begin{figure}[t!]
    \centering
    \includegraphics[width = 0.48\textwidth, trim = 0 0 0 0, clip]{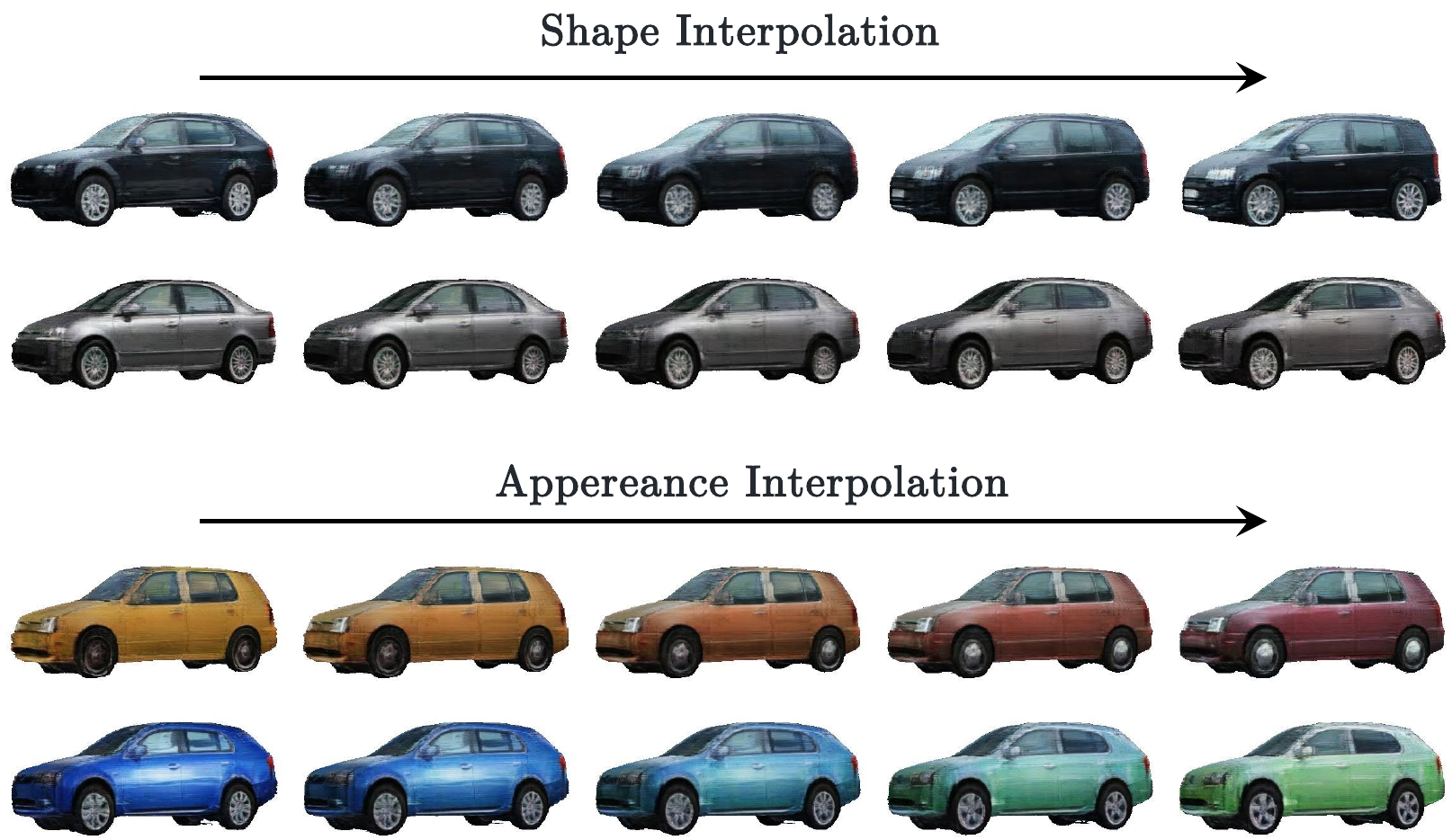}
    \caption{Visualization of interpolation. We linearly interpolate the latent codes of NeRF from left to right. Our method faithfully generates a smooth and meaningful transition from one to another, in both shape and appearance.}
    \label{fig:interpolation}
\end{figure}

\begin{table}[b!]
    \centering
    {
    \addtolength{\tabcolsep}{1pt}
    \centering

    \begin{tabular}{|l|c|c|c|c|c|c|}
    \hline
    \textbf{Method} &  {Easy} &  {Mod} &  {Hard}  \\
    \hline
    Original data & 21.57  & 15.51 & 13.58 \\
    ShapeNet model & 19.64 & 15.75 & 13.80 \\

    \textcolor[RGB]{100,100,100}{GIRAFFE~\cite{Niemeyer2020GIRAFFE}} & \textcolor[RGB]{100,100,100}{20.79} & \textcolor[RGB]{100,100,100}{15.10} & \textcolor[RGB]{100,100,100}{12.50} \\
    
    \textcolor[RGB]{100,100,100}{GIRAFFE HD~\cite{xue2022giraffehd}} & \textcolor[RGB]{100,100,100}{19.30} & \textcolor[RGB]{100,100,100}{14.44} & \textcolor[RGB]{100,100,100}{12.39} \\

    SimpleCopyPaste~\cite{lian2022copypaste} & 22.23 & 15.47 & 13.24 \\
    
    Ours & \textbf{24.07} & \textbf{18.09} & \textbf{15.06} \\
    \hline
    \end{tabular}%
    }

    \vspace{0.5mm}
    \caption{Quantitative Evaluation on the KITTI validation dataset. $AP|_{40}$ of 3D bounding box on the Car category are reported. 
    }
    \label{tab:kitti_exp}
\end{table}

\myparagraph{Quantitative Results}
\quad In Tab.~\ref{tab:kitti_exp}, we provide the experimental results of leveraging the augmented datasets to train 3D object detectors. Following existing settings~\cite{lian2022copypaste, wang2022neural}, we use CenterNet~\cite{zhou2019objects} (KITTI) and FCOS3D~\cite{wang2021fcos3d} (nuScenes) trained on original data as baseline detectors. We compare our method with several methods: \textbf{ShapeNet model} denotes that we randomly select 20 CAD models from ShapeNet car, rendering the models under different positions and rotations to generate an augmented dataset. \textbf{GIRAFFE} and \textbf{GIRAFFE HD}: Since they don't contain real-world scale and can not generalize to new camera parameters, we only paste their generated objects in the center of images and leverage object height to recover the ground truth depth. We lighten their color for such issues. \textbf{SimpleCopyPaste} is SOTA copy-paste augmentation method in monocular 3D detection. As discussed in~\cite{lian2022copypaste}, it is restricted to using existing objects with fixed viewpoints to do augmentation, while our method can generate novel objects in arbitrary positions and rotations. Consequently, our method can effectively enhance the diversity of training data and achieve better results as shown in Tab.~\ref{tab:kitti_exp}.



\begin{figure}[t!]
    \centering
    \includegraphics[width = 0.48\textwidth, trim = 0 0 0 0, clip]{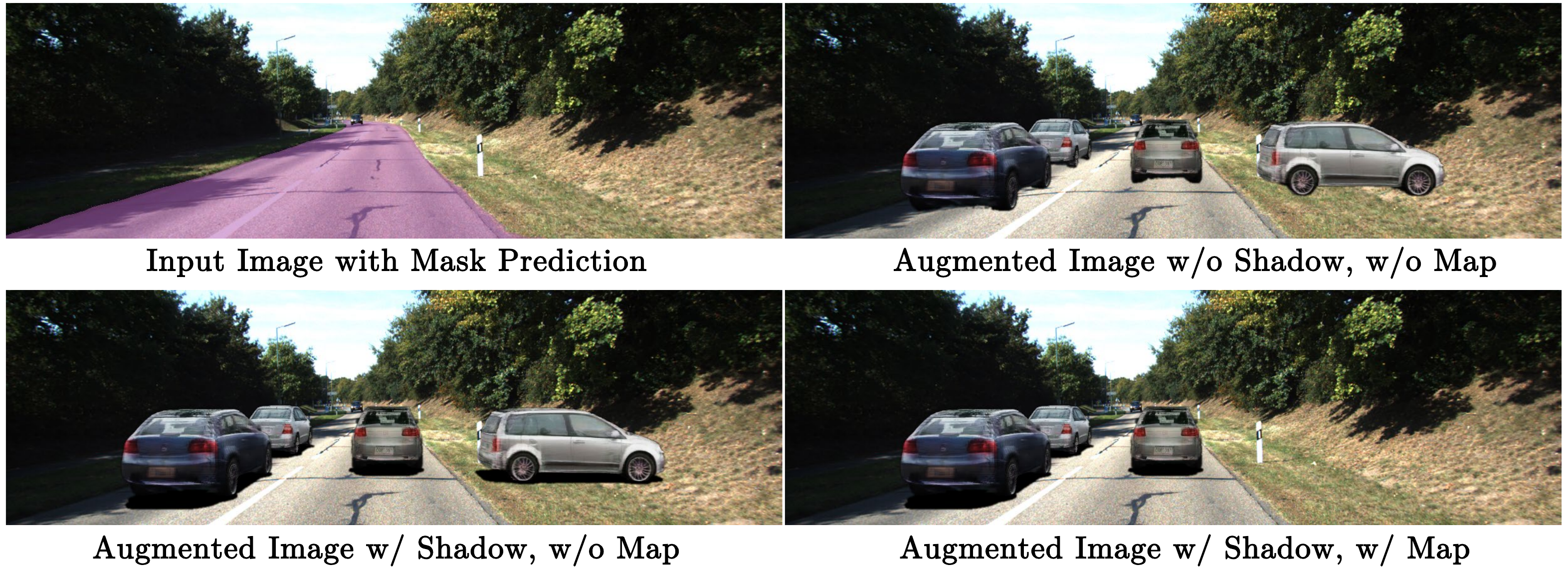}
    \caption{Visualization of different composition strategies. ``Map'' denotes only sample objects in drivable areas. ``Shadow'' denotes whether to add a shadow at the bottom of objects.}
    \label{fig:abla_map_shadow}
\end{figure}

Except for the KITTI dataset, we also validate the effectiveness of our method on the nuScenes dataset. We augment the full-surround view images of dataset with our method. We use the same training strategy and model hyperparameters as~\cite{wang2021fcos3d} but do not supervise attributes and velocity as they are not included in the augmented data. It is worth mentioning that, by only augmenting the car objects, we can achieve consistent gain among the majority of classes even though we do not directly augment these classes.

\begin{table}[h!]
    \centering
    \resizebox{1.0\linewidth}{!}{
    \addtolength{\tabcolsep}{1pt}
    \centering
    \begin{tabular}{|l|c|c|c|c|c|c|}
    \hline
    \textbf{Method} &  {mAP} &  {Car} &  {Bus} & {Pedestrian} &  {Motorcycle} \\
    \hline
    Original data & 0.136  & 0.231 & 0.072 & 0.277 & 0.142  \\
    Ours & \textbf{0.146} & \textbf{0.254} & \textbf{0.086} & \textbf{0.297} & \textbf{0.154} \\
    \hline
    \end{tabular}%
    }
    \vspace{0.5mm}
    \caption{
    Experimental results of FCOS3D~\cite{wang2021fcos3d} on the nuScenes validation set. mAP represents the mean average precision of 10 object categories. We additionally present the AP of other classes that are not directly augmented (\textit{e.g.}, bus, pedestrian, and motorcycle). 
    }
    \label{tab:nus_exp}
\end{table}

\begin{table}[b!]
    \centering
    {
    \addtolength{\tabcolsep}{1pt}
    \centering

    \begin{tabular}{|l|c|c|c|c|c|c|}
    \hline
    \textbf{Method} &  {Easy} &  {Mod} &  {Hard}  \\
    \hline
    SDF-Label~\cite{sdflabel} & 1.23  & 0.54 & N/A \\
    Our unsup. data & \textbf{5.15} & \textbf{3.80} & 3.48 \\
    \hline
    \end{tabular}%
    }

    \vspace{0.5mm}
    \caption{Comparison with unsupervised 3D detection methods. Note that SDF-Label~\cite{sdflabel} additionally uses LiDAR point clouds for accurate depth retrieval.
    }
    \label{tab:unsup}
\end{table}

\myparagraph{Unsupervised 3D Object Detection}
\quad Besides augmenting existing datasets, our framework can create datasets from scratch, which means unsupervisedly train a 3D detector. Specifically, we select the images with more than 90\% of background pixels as empty images by using a pretrained segmentation network~\cite{SunXLW19}. Then we populate novel objects into these nearly-empty background images. During the whole process, the construction of the training data does not require any human annotation for 3D bounding boxes. As shown in Tab.~\ref{tab:unsup}, our trained model outperforms existing unsupervised training solution by a large margin.

\subsection{Ablation Study}\label{sec:ablate}

We ablate different composition strategies that affect performance in the downstream task in Tab.~\ref{tab:ablate_kitti}. We fix the sampling pose of all synthesized objects and train the same model~\cite{zhou2019objects} with different composition strategies. As shown in Fig.~\ref{fig:abla_map_shadow} and Tab.~\ref{tab:ablate_kitti}, the improvement of downstream task performance is consistent with the enhanced visual quality. These indicate the importance of realistic simulation in downstream tasks.


\begin{table}[htbp]
    \centering

        \begin{tabular}{|c|c||c|c|c|c|} \hline
         Map & Shadow & Easy & Mod & Hard \\ \hline
             &    &    22.35   &  17.36     &14.53  \\ 
         \cmark      &            &   22.85    &  16.49 & 14.08  \\
              &    \cmark      &   23.08    &  17.71 & 15.02  \\
         \cmark  &\cmark  & \textbf{24.07} & \textbf{18.09} & \textbf{15.06} \\
         \hline
        \end{tabular}%

        \caption{Ablation of different composition strategies on the KITTI validation dataset. ``Map'' denotes road layout aware sampling. ``Shadow'' denotes whether to cast a shadow at the bottom of objects.}

    \label{tab:ablate_kitti}

\end{table}%

\section{Limitation and Future Work}\label{sec:limitation}

\myparagraph{Relation Reasoning}
\quad The proposed method augments the existing dataset by sampling novel objects uniformly on the ground plane. If objects are not properly sampled in reasonable positions, they can cause unharmonious in the augmented images. For example, a vehicle may not follow the lane lines or even rush to a building. Also, relations between traffic participants are not modeled realistically. In the future, these issues could be addressed by traffic flow simulation~\cite{suo2021trafficsim, tan2021scenegen}, or geometry-aware composition~\cite{chen2021geosim}.

\myparagraph{Photorealistic Appearance}
\quad Since our generated objects have a different data source from the original datasets. There exist an illumination gap between generated object and the existing environment. For example, the rendered object in driving scene should be darker in the afternoon, compared with midnoon. Lighting estimation~\cite{wang2022neural, yao2022neilf} can help with this problem. In the meantime, high-quality material properties should be estimated in detail. Techniques proposed in GET3D~\cite{gao2022get3d} could potentially be used to generate objects with high-fidelity textures.

\section{Conclusion}
\label{sec:conclusion}


We propose Lift3D, a 3D generation framework that provides high-resolution synthesis and tight 3D annotation, with the goal of training data generation. Compared with other 3D generative models relying on a 2D upsampler, our lifting process gives explicit 3D information of generated objects and hence the objects align well with given 3D labels. Also, our approach provides more realistic appearance generation by distilling the knowledge from pretrained StyleGAN2. Based on the proposed framework, we can enhance original datasets both in quantity and diversity. Experimental results show significant improvements over baseline in monocular 3D object detection tasks. We hope this work can take us a step further in 3D asset generation for driving scene simulation.




{\small
\bibliographystyle{ieee_fullname}
\bibliography{11_references}
}



\end{document}